\RequirePackage{fix-cm}
\documentclass[smallextended]{svjour3}       
\smartqed  
\usepackage{graphicx}
\usepackage{booktabs}
\usepackage{natbib}
\usepackage{algorithmic}
\usepackage{tabularx}
\usepackage{booktabs}
\usepackage{url}

\begin{document}

\title{A Heuristic Method for Solving the Problem of Partitioning Graphs with Supply and Demand}

\titlerunning{Partitioning Graphs with Supply and Demand}        

\author{Raka Jovanovic    \and     Abdelkader Bousselham \and
        Stefan Vo{\ss} 
}


\institute{Raka Jovanovic \at
 Institute of Physics, University of Belgrade, Pregrevica 118, Zemun, Serbia\\
              \email{rakabog@yahoo.com} \\          
             \emph{Present address:} Qatar Environment and Energy Research Institute (QEERI),
  PO Box 5825, Doha, Qatar 
		\and
		Abdelkader Bousselham \at
		 Qatar Environment and Energy Research Institute (QEERI),
  PO Box 5825, Doha, Qatar
              \email{abousselham@qf.org.qa}           
		\and
         Stefan Vo{\ss} \at
         Institute of Information Systems, University of Hamburg, Von-Melle-Park 5, 20146 Hamburg, Germany \email{stefan.voss@uni-hamburg.de}
}

\date{Received: date / Accepted: date}

\maketitle

\begin{abstract}
In this paper we present a greedy algorithm for solving the problem of the  maximum partitioning of graphs with supply and demand (MPGSD). The goal of the method is to solve the MPGSD for large graphs in a reasonable time limit. This is done by using a two stage greedy algorithm, with two corresponding types of heuristics. The solutions acquired in this way are improved by applying a computationally inexpensive, hill climbing like, greedy correction procedure. In our numeric experiments we analyze  different heuristic functions for each stage of the greedy algorithm, and  show that their performance is highly dependent on the properties of the specific instance.  Our tests show that by exploring a relatively small number of solutions generated by  combining different heuristic functions, and applying the proposed correction procedure we can find solutions within only a few percent of the optimal ones.
\keywords{Graph Partitioning  \and Greedy Algorithm  \and  Demand vertex \and Supply vertex ·}
\end{abstract}

\section{Introduction}
\label{intro}

A wide range of practical problems can be efficiently represented by means of graph partitioning. Due to this, many variations of the problem have been analyzed which correspond to specific applications like having a balanced partitioning \citep{Andreev:2004:BGP:1007912.1007931}, minimizing the number or weight of cuts \citep{Reinelt2008385,doi:10.1137/0401030}, or by limiting the number of cuts \citep{Reinelt20101}.  In this paper the focus is on the problem of {\it maximum partitioning of a graph with supply and demand} (MPGSD). This problem is defined on a graph $G$, in which each node is either a supply or a demand node. Each vertex $v$ has a corresponding positive number, which is called the supply of node $v$; otherwise, if $v$ is a demand node, this value would be called demand.
Each demand node can at most receive power from one supply node through the edges of $G$. The goal is to partition $G$ into connected subgraphs $S$, by deleting
edges in $G$ in a way that each of them has exactly one supply node, whose supply is at least the sum of demands of all demand nodes in $S$. The goal is to maximize the total demand that
is included in the subgraphs.
One of the important applications of this theoretical graph problem is modeling of  self-adequacy of interconnected micro-grids \citep{SelAdeq}, although in a much simplified form.  The  maximum partitioning problem has been applied for several problems in the field of  power supply and  delivery networks \citep{Simp1, Tree1, Simp2, Simp3}.

In existing literature the main focus of research for the MPGSD has been on theoretical aspects of the problem \citep{Ito2008627,Tree2,Tree1,Tree3}.
One direction of this research was in developing methods to create relaxed versions of the problem. Due to the complexity of the problem, the research up to now has focused on some specific type of graphs like trees \citep{Tree2, Tree1,Tree3} and series-parallel graphs \citep{Ito2008627}. An interesting approach for solving this problem, for general graphs, is given in \cite{SupDem}. In it an algorithm  based on the calculation of all the paths from supply nodes to demand nodes, has been developed for applications in power supply networks. Due to the fact that all such paths need to be calculated/updated the algorithm is computationally expensive. The proposed algorithm gives a guarantee of a $2k$ approximation, where $k$ is the number of supply nodes.  The original problem has also been extended with additional properties to better suite some specific applications. One such example is the  parametric version of the problem, in which all  the demands and supplies are dependent of a single parameter $\lambda$ for application to power supply networks \citep{Parametric}.  Another extension is the problem of minimum cost partitions  of  trees with supply and demand in which a maximal capacity has been added to edges \citep{Tree4}.

It should be noted that the MPGSD at first glance looks similar to the capacitated $p$-median problem which seeks to choose $p$ nodes as median nodes and then assigns all the remaining nodes to one of those medians; see, e.g., \cite{Bionomic}. In the capacitated version of the problem the medians have a limited capacity but any node may be assigned to any median if the capacity constraint allows this (in an idealized form assuming the existence of a complete graph). For the MPGSD this is not as easy as we need to find disjoint subgraphs connected to the (let us call them) medians observing the capacity constraint.
In a similar way we may argue when it comes to the capacitated clustering problem; see, e.g., \cite{Scheuerer2006533}.
Here a given set of nodes with known demands has to be partitioned into $p$ distinct clusters.
Each cluster is made up by a customer acting as its cluster center. The objective is to minimize the sum of distances from all cluster centers to all other customers in their cluster, such that a predefined capacity limit of the cluster is not exceeded and every customer
is uniquely assigned to one cluster.
As a final example for other types of somewhat related partitioning problems consider the partitioning of the vertices of a given graph into a number of balanced classes, while minimizing the number of cut edges between the classes \citep{Galinier2011}.

In this paper we present a greedy heuristic approach to solving the MPGSD.  The goal of developing such an algorithm is to allow solving large scale instances, which is needed when modeling real power systems especially in the case of interconnected microgrids. It has previously been shown that heuristics are very efficient in the case of other types of partitioning problems \citep{doi:10.1137/0401030,BLTJ:BLTJ1770,Kim:2004:LBG:974494.974496}. It is important to point out that in the existing research there is a lack of experimental results and benchmark data sets. With this work we aim to fill this gap by providing benchmarks and making the code developed for the proposed method available for download, that could be used as a reference for future research.

The proposed method is based on a two stage greedy algorithm. Further, two greedy correction procedures are developed to improve the quality of acquired solutions. The paper also includes an analysis of several heuristic functions that are used in the two stages of the algorithm. We wish to emphasize that the goal of this research was to develop a method that can find good approximate solutions  within reasonable computational time. Because of this the developed algorithm does not incorporate a look ahead mechanism based on the analysis of adding more than one node to the partial solutions at each iteration. Although this type of lookahead frequently significantly improves the quality of solutions acquired by greedy algorithms, this is achieved at a significant increase of the calculation time (unless some related finetuning is performed).

In our numeric experiments we have observed that although certain heuristics have had an overall better performance, in the sense of having the best average of found solutions, there would be a wide discrepancy when individual problem instances are observed. Such behavior would be even more evident when the correction procedure was included.  It is a well known fact that when having several competing heuristics for an optimization problem, it is good practice to apply all of them and simply select the best found solution. This idea can effectively be applied for the MPGSD, since we can combine different heuristics for each of the stages of the algorithm. In this way we can explore a relatively small number of potentially good solutions. In our test we show that such an approach can find approximate solutions that have an average error of 1-2\% when trees or general graphs are considered.	
	
The paper is organized as follows. In Section~\ref{SupplyDemandSection2} we define the MPGSD more comprehensively. Then we present the proposed algorithm, with separate subsections dedicated to each of the stages of the algorithm and the correction procedure. In Section~\ref{MPGSDResults} we show the results of our computational experiments.

\section{Maximal Partitioning  of a Graph With Supply/Demand}\label{SupplyDemandSection2}

In this section we give a definition for the MPGSD as a slightly modified version of the one given in  \cite{Ito2008627}.


The MPGSD is defined for an undirected graph  $G=(V,E)$  with a set of nodes $V$ and a set of edges $E$. The set of nodes $V$ is split into two disjunct subsets $V_s$  and $V_d$. Each node $u\in V_s$, is called a supply vertex and will have a corresponding positive integer value $sup(u)$. Elements of the second subset $v \in V_d$ are called demand vertices and  will have a corresponding positive integer value $sup(v)$.   The goal is to find a partition $\Pi= \{S_1, S_2,.., S_n\}$ of the graph $G$ that satisfies the following constraints.  All the subgraphs in $\Pi$ must be connected subgraphs  containing  only  a single distinct supply node. As a result we have $|V_s| = n$.  Each of the $S_i$  must have a supply greater or equal  to its total demand. Each demand vertex can be an element of only one subgraph, or in other words it can only receive `power' from one supply vertex through the edges of $G$.

With the intention of having a  simpler notation, we will use strictly  negative values for demand nodes and positive values for supply nodes.  The goal is to maximize the fulfillment of demands, or more precisely to maximize the following sum.
\begin{equation}
- \sum_{S \in \Pi}\sum_{v \in S\cap V_d}sup(v)
\end{equation}
while the following constraints are satisfied for all $S_i \in \Pi$
\begin{eqnarray}
\label{constarint}
\sum_{v \in S_i}sup(v) \geq 0 \\
\label{constarint2}
S_i \cap S_j = \emptyset  \,\,\,\,\, ,i\neq j\\
S_i \,\,\,\, is \,\,\,\,connected
\end{eqnarray}

\begin{figure*}[tcb]
\centering
\includegraphics[width=0.9\textwidth]{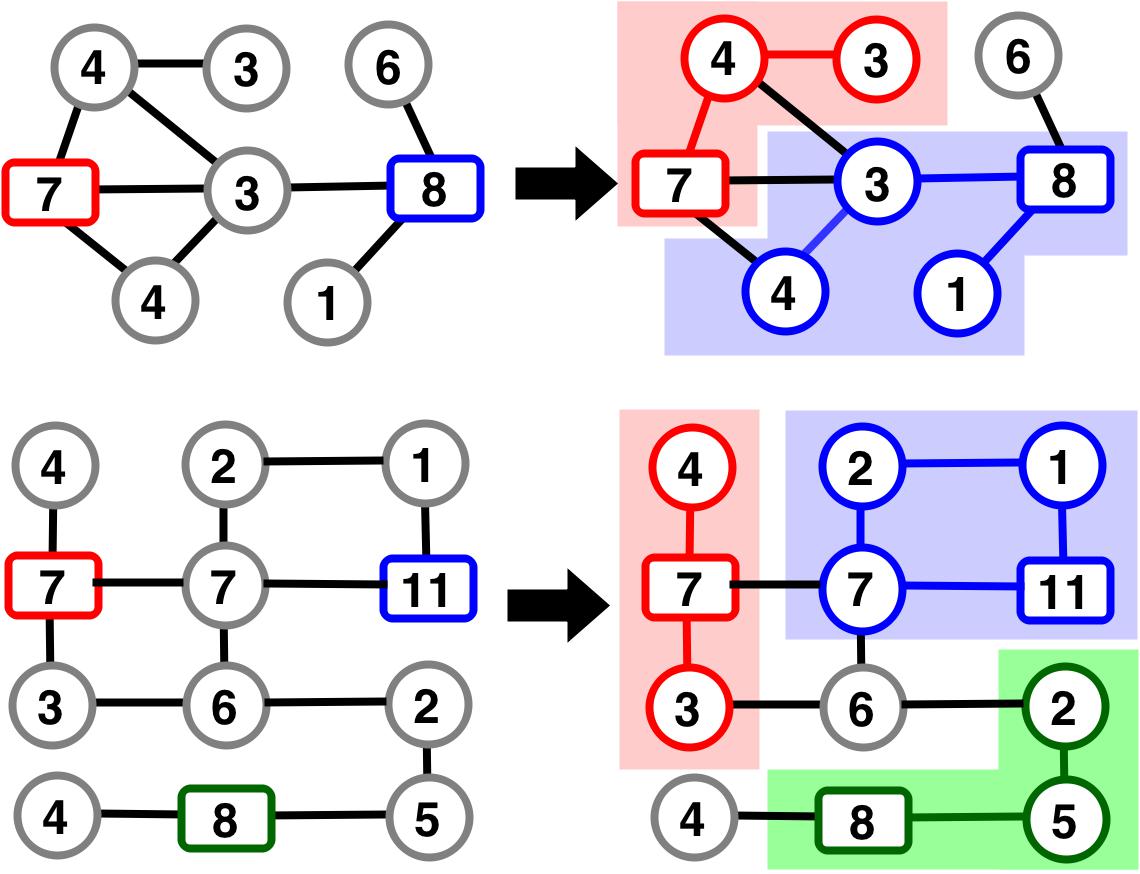}
\caption{Examples of problem instances for the MPGSD. On the left the  square nodes represent supply nodes and circles demand nodes. Numbers within the nodes correspond to supply and demand values, respectively. The right side shows the solutions, where the same color of nodes indicates they are a part of the same partition.}
\label{fig:Problem}
\end{figure*}

 It has been shown that the MPGSD is NP-hard even in the case of a graph containing  only one supply node and having a star structure \citep{Ito2008627}. An illustration of problem instances and corresponding solutions for the MPGSD is given in Figure \ref{fig:Problem}.

\section{Outline of the Algorithm}

In this section we will first give a short outline of the greedy algorithm and corresponding corrections and in the following section each of the steps will be presented in detail. As previously mentioned the solution of the MPGSD will consist of $|\Pi|=n$ subgraphs, where $n=|V_s|$ is the number of supply nodes.  The idea of the greedy algorithm is to start with $n$ disjunct subgraphs $S_i$. At the initial step of the algorithm each of  the subgraphs would consist of  only one supply node $s_i$. Next,  at each step (iteration)  we would expand one of the subgraphs $S_i$ with one vertex $v \in V_d$. In practice each iteration would consist of two sub-stages. In the first we would select which subgraph $S_i$ would be expanded, and in the second stage we would decide by which node $v$.  The selection of $v$  will be constrained in a way that the newly generated subgraph is connected, and that  $v$ is not an element of any other subgraph. This selection should also produce a subgraph whose total demand is equal or less to its  supply.   Of course, the selection of $S_i$ and $v$ should be done using some heuristic measure that we expect will produce high quality solutions of the problem.

It is common practice to improve the solutions  acquired using direct greedy algorithms based on some local search to explore similar ones within a specific neighborhood. Examples of such an approach are the 2-opt and 3-opt algorithms with their application to problems that can formally define suitable neighborhoods. The problem with such approaches is that if we chose a small neighborhood the chances of improvement are limited. On the other hand if the chosen neighborhood is large the method becomes computationally expensive, in many case even similar to the use of some advanced meta-heuristic that can produce approximate solution of higher quality. This is especially a problem with large scale test instances, where even if the neighborhood is defined relatively strictly, it will still consist of a large number of potential candidates. As it is discussed in  \cite{doi:10.1137/0401030}, in case of graph partitioning problems this approach generally equates to exchanging nodes between subgraphs.

In our methods, we propose using a greedy correction approach that can find improvements of the original solution by exploring only neighboring solutions that are selected using a heuristic procedure. Two such greedy correction procedures are used. In the first we simply check if some node $v$, that is not an element of any of the created subgraphs, can be exchanged with  some node $u$ of a neighboring subgraph $S_i$ to increase the total covered demand. As it is mentioned in \cite{SupDem}, which we have also experienced in our initial work on the MPGSD \citep{BasicAlgorithm}, one of the main drawbacks of greedy algorithms is that a subgraph can be easily cut off from the rest of the graph as nodes are added to other subgraphs. As a consequence of this, it is possible that some supply nodes will be covering only a very small amount of demands. The second correction procedure attempts to resolve this problem.

\subsection{Greedy Algorithm}

To formally define our algorithm, first we shall specify a corresponding set  $NV$  to $v \in V$
where $NV(v)$ represents the set of adjacent nodes to $v$ in $G$:
\begin{equation}
	NV(v) = \{u| u \in V \wedge (u,v) \in E\}
\end{equation}

The idea is to slowly grow each of the subgraphs $S_i$ at each iteration by adding new nodes. Since the subgraphs will be changing at each step of the algorithm we introduce the notation $S_i^k$ for the state of subgraph $S_i$ at iteration $k$. We will define the function $NV^k$ for subgraphs  $S_i$, that represents the set of all the nodes that are connected to some node in $S_i^k$.
\begin{equation}
    \label{NV_SG}
	\hat{N_i^k} = NV^k(S_i) = \{u| u \in V \wedge \exists (v \in S_i^k) (u,v) \in E\}
\end{equation}
It is obvious that if at some step $k$ we select some node $v \in \hat{N_i^k}$, and adding it to $S_i^k$, the resulting subgraph $S_i^{k+1}$ will be connected. The selection of $v$ must also be defined in a way that the constraints given by  Equations (\ref{constarint}), (\ref{constarint2}) are satisfied,  or in other words that there is no such $S_j^{k+1}$ for which $v\in S_j^{k+1}$ and that the demand has not become larger than the supply in $S_j^{k+1}$.  To achieve this, we define a corrected set of nodes $N_i^k$, such that by adding $v \in N_i^k$ to $S_i^k$ the new subgraph will satisfy all the constraints. To do this we first define function $sup_i^k$ as the available supply for each $S_i^k$ given in Equation (\ref{Available}).

\begin{equation}
    \label{Available}
	 sup_i^k = \sum_{v \in S_i^k}sup(v)
\end{equation}

Now we can define $N_i^k$ in the following way.
\begin{equation}
    \label{NV_SG2nd}
	N_i^k = \{u| u \in \hat{N_i^k} \wedge |sup(u)|\leq sup_i^k\} \setminus \bigcup^n_{j=1}S_j^k
\end{equation}

It is important to mention that the sets of neighbors $N_i^k$ can be efficiently calculated using an auxiliary structure that helps track the used nodes by updating the set of edges $E$ for graph $G$, using a similar procedure like in the case of covering problems given in \cite{Jovanovic20115360,Raka2}. Details of the implementation are presented in our previous work  \citep{BasicAlgorithm}.

Using the sets $N_i^k$, we can define a greedy algorithm for the MPGSD using two heuristic functions. At each step of the algorithm, the first heuristic $hs$ is used to select the best $S_i$, and the second heuristic $hv$ will be used to select the best $v \in N_i^k$ to add to $S_i$. The algorithm stops if all the neighboring sets $N_j^k$ are empty. In the following two subsections, we present the types of heuristic functions  used at each stage of the greedy algorithm.

\subsubsection{Heuristics for Subgraph Selection}

We first define a  heuristic function $hs$ that gives us the  desirability of  selecting a subgraph $S_i^k$ at iteration $k$ for  expansion. There are two main properties of $S_i^k$ that should be
considered: the  available supply $sup_i^k$, or in other word the amount of supply that has not been covered by some demand, and the number of potential candidates $|N_i^k|$.

One heuristic will consider subgraphs with a higher value of $sup_i^k$  as more desirable, in the sense that they should be selected earlier. The logic behind this heuristic is  that it is expected that more nodes need to be added to such subgraphs than to ones with a lower available supply. In the proposed greedy algorithm  the number of non-satisfied demand nodes is constantly decreasing and as a consequence there exists a possibility  that there may not be enough available demand to reach $sup_i^k$. Using this idea we define the heuristic $hs_1$ as
\begin{equation}
    \label{HS1}
	hs_1(S_i^k) = sup_i^k
\end{equation}

As mentioned before, in the proposed algorithm we can enter a state in which it is not possible to expand subgraph $S_i^k$ due to being cutoff from the rest of the graph by other subgraphs. Since  each node can be an element of only one subgraph, there is a possibility that all the elements of $N_i^k$  will be added to other subgraphs since a node can be a neighbor of several different subgraphs. Because of this we shall consider subgraphs with a low value of $|N_i^k|$ highly desirable. The second heuristic $hs_2$ can be formally defined as
\begin{equation}
    \label{HS2}
	hs_2(S_i^k) = \frac{1}{|N_i^k|} \,\,\,\, ,|N_i^k|\neq 0\\
\end{equation}
Finally we introduce a third heuristic function $hs_3$ that balances the two effects
\begin{equation}
    \label{HS3}
	hs_3(S_i^k) = \frac{sup_i^k}{|N_i^k|} \,\,\,\,   ,|N_i^k|\neq 0\\
\end{equation}

We should point out that for the heuristic functions $hs_2$ and $hs_3$ the case $|N_i^k|=0$ will be considered least desirable and will never be selected.

\subsubsection{Heuristics for Node Selection}

The second type of heuristic function $hn$ that needs to be defined is used    for giving the desirability of adding $v \in N_i^k$ to $S_i^k$. In designing such heuristic we shall follow a similar logic as in the case of heuristic functions for subgraphs.

Let us first define function $hn1$ as
\begin{equation}
    \label{HV}
	hn_1(v) = |sup(v)|
\end{equation}

In Equation (\ref{HV}) nodes $v$ with high demand are considered more desirable. The idea of this heuristic is that it  gets harder to satisfy high demands as the algorithm progresses due to fact that the available supply constantly decreases as new nodes are added to the subgraph. Because of this it seems better to resolve high demands early.

The second heuristic for selection of nodes is designed to avoid the problem of subgraphs being cutoff from the rest of the graph. In this case  the desirability of node $u \in N_i^k$ will be proportional to  increases/decrease of the number of elements of $N_i^{k+1}$. Formally we can define the new heuristic $hn2$ as
\begin{equation}
     \label{ExtensionN}
     C(u,S_i^k) = \{v| v \in NV(u) \wedge \neg (v |\in {N_i^k}) \wedge (sup(v) \leq sup_i^k+sup(u))\}
\end{equation}
\begin{equation}
     \label{HN2}
     hn_2(u) =  |C(u,S_i^k)|
\end{equation}
$C(u,S_i^k)$ is defined as the set of new neighbors that will be added. This set consists of neighboring nodes to $u$, which are not already in $N_i^k$ and whose demand is not greater than the available supply of $S_i^k$ after adding node $u$. The heuristic function $hn2$ is now simply the number of elements of $C(u,S_i^k)$   as given in Equation (\ref{HN2}).

As in the case of selecting a subgraph, in case of nodes we will also use a balanced heuristic. In this case nodes with a high demand and a significant increase in the number of neighbors of the corresponding subgraph are considered highly desirable. We define a new heuristic.
\begin{equation}
     \label{HN3}
     hn_3(u) =  (hn_2(u) +1) hn_1(u)
\end{equation}
In Equation (\ref{HN3}) the addition of one is included to distinguish between nodes that do not add new connections to the subgraph.

\subsubsection{Approximate computational cost}

\begin{figure*}[tcb]
\centering
\includegraphics[width=0.9\textwidth]{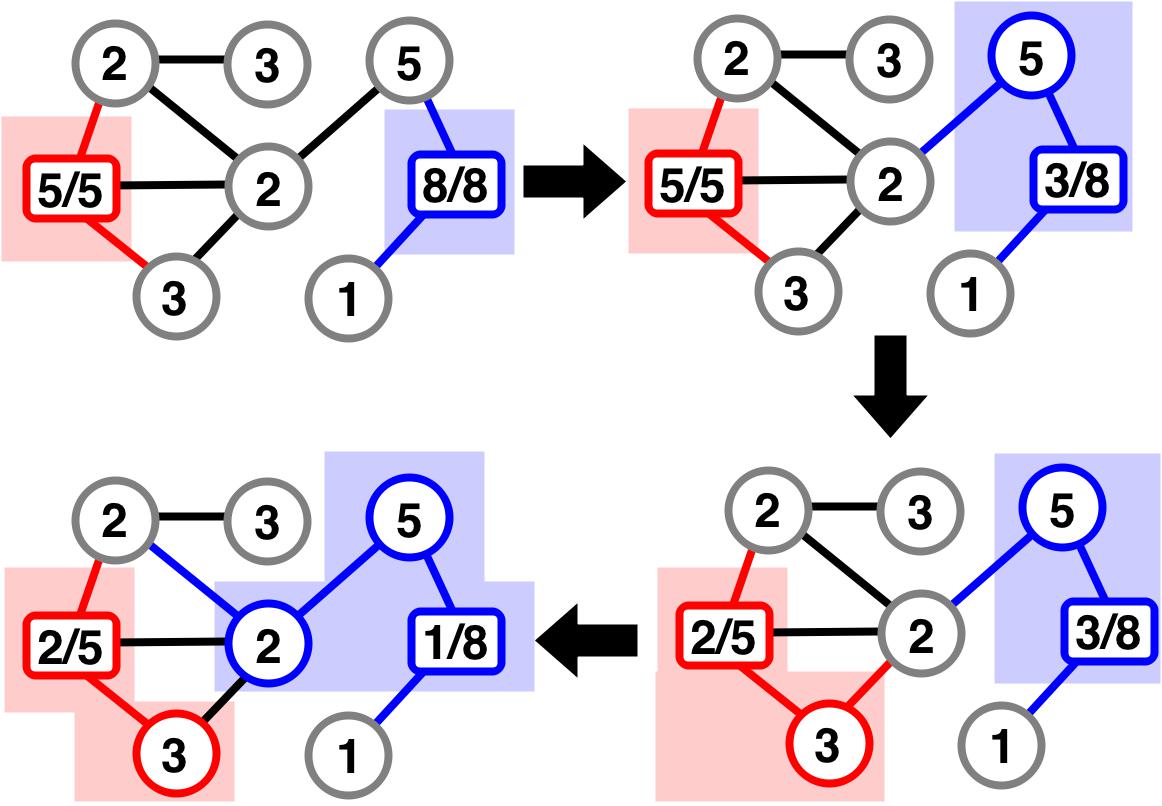}
\caption{Illustration of several steps of the greedy algorithm when heuristics $hs1$ and $hn1$ are used. The second number for supply nodes indicates the number of available supply. The colored edges connected to grey(non-located nodes) give us the neighborhood of a partition.  }
\label{fig:Greedy}
\end{figure*}

The proposed greedy algorithm can be presented using the following pseudo code (see also Figure~\ref{fig:Greedy} for the illustration of several steps of the algorithm):

\begin{algorithmic}
 \STATE
\STATE Initialize All $S_i$ with supply nodes; choose appropriate heuristic functions $hs$ and $hn$
 \WHILE{$(Sum(sup_i)>0)$ \textbf{and} $Sum(|N_i|)>0)$}
	\STATE {Select $S_i$ using $hs(S_i)$}
	\STATE{Select $u \in N_i$ using $hn(u)$}
	\STATE{Add $u$ to $S_i$}
	\STATE{Perform Updates}
 \ENDWHILE
\end{algorithmic}

In the algorithm we first initialize all of the subgraphs as different supply nodes.
The iterative algorithm is repeated until all the available supply is covered or no subgraph has any neighboring nodes to which it can expand.  We first select subgraph $S_i$ that is to be expanded and then the node with which it will be expanded. Finally, the necessary updates to auxiliary structures are performed. In our previous work \citep{BasicAlgorithm} we give detailed explanation of the auxiliary structures and how they are used for a simpler version of the algorithm.

Our goal of this work is more dedicated towards experimental investigation rather than theoretical analysis. Therefore, we only present a descriptive approximation of the asymptotic calculation time of the proposed algorithm. This time can be approximated as follows:
\begin{equation}
     \label{CalcGreedy}
     |V_d|*(|V_s|+|AvgNumNeighbors(S_i)|+AvgNumConnections(u))
\end{equation}

In Equation (\ref{CalcGreedy}) the term $|V_d|$ is included as we will be adding demand nodes one by one to the partial solution. At each iteration we will first select the most desirable subgraph or, in other words, finding the maximal value of $|hs|$ for which we will need $|V_s|$ operations. It is important to mention that this can be done more efficiently, since at each step only a limited number of subgraphs will be changed. For the selection of the best node for expansion, we will have to check all the neighbors of a subgraph $S_i$ which gives us the second term $|AvgNumNeighbors(S_i)|$. This value will be highly dependent on the density of the graph, but will never be higher than $|V_d|$. This has the consequence that the proposed algorithm will be much more efficient in case of sparse graphs. In our experiments we have seen that this value is generally significantly lower than $|V_d|/|V_s|$. The proposed heuristic functions can be calculated in constant time if suitable auxiliary structures are used. The calculation time for the update procedure for such structures will be proportional to the number of connections that   the added node $u$ has.

\subsection{Greedy Correction}

In this section we present a greedy correction method for improving the solutions acquired using the algorithm presented above. As mentioned before, in this way we wish to avoid testing a large neighborhood of  solutions acquired by the greedy algorithm. This shall be done in two ways, in the first correction we shall attempt to incorporate some of the demand nodes that have not yet been included into the generated subgraphs. This is done by exchanging them with some node of the generated subgraphs. The second correction focuses on the problem that some of the supply nodes, and corresponding subgraphs, can be cut off from the rest of the graph at a very early stage of the algorithm. Note that this case frequently results with partitioning in which a large amount of demand has not been covered. The idea of this correction is to expand such subgraphs at the cost of their neighbors with the hope that the amount of the covered demand can be increased by applying the first correction procedure. For both types of corrections it is essential that the connectivity of the corrected subgraphs is maintained. As it is well known, testing the connectivity of a subgraph can be a computationally expensive procedure so we shall first introduce a simple auxiliary structure that can be used to simplify this test.

\subsubsection{Auxiliary tree structure}

There have been several algorithms developed for finding all the articulation points \citep{Chaudhuri:1998:ODA:2247768.2248151,ArtPoint}, i.e., nodes whose removal will result in an unconnected graph, but such a calculation would be redundant  for the needs of the proposed algorithm. In the proposed method we only wish to avoid selecting an articulation point, in case it satisfies some other constraints. Because of this we will use an approximate approach for recognizing such points to avoid unnecessary calculations.

The idea is to represent each of the subgraphs as a tree. In this way only leafs will be considered in the correction procedure. It is not necessary to have a complete tree structure but only a method that makes it possible to track leafs. Such a structure should be simple to update when nodes are added/removed from the subgraph, and is preferred to have a higher number of leafs. This can be done  in the following way, each node of the graph will have a corresponding integer value $ch(u)$ indicating the number  child nodes in the tree. We can generate such a structure using the following simple procedure when generating the partitioning. Let us consider the case that node $u$ is added/removed with respect to subgraph $S_i^k$

\begin{enumerate}
\item{ Each node $u$ has a corresponding value $ch(u)$ which indicates the number of children it has in a tree. Initially all nodes have $ch(u)=0$, except  supply nodes which have the maximal possible value of $ch$, since they can not be removed from a subgraph}
\item{ Select all the neighbors of $u$ that are elements of $S_i^k$}
\item{For a node $v$ among the nodes which have the largest value $ch(v)$, increment $ch(v)$. In case of several nodes with the same value $ch$ simply select the one with a lower $id$ (number of the node)}
\item{When a leaf node $u$ is removed form $S_i^k$, select from its neighbors $v \in S_i^k$ the one with the largest value of $ch(v)$, and decrease $ch(v)$. In case of several nodes having the same value of $ch$ select the one with a lower $id$}

\end{enumerate}

Using this structure it is possible to check if a node is a leaf in constant time. The update procedure for maintaining this structure only depends on the number of connections that a node has. It is important that this structure gives in some cases false positives, for articulation points. From our experience this would very seldom  effect the performance of the algorithm.

\subsubsection{Correction for Non-located Nodes}

The idea behind the first corrections is that it is possible to exchange a node $u$ that is non-located, with some leaf node $v$ of a subgraph and increase the covered supply. Here we use the term non-located for nodes that are not a part of any of the generated subgraphs. It is expected that this type of procedure will not be computationally expensive due to the fact that only a low number of nodes will be left non-located after the greedy algorithm has been applied.

To fully specify this type of correction procedure, we first formally define the set of nodes $v \in S_i$ that can be exchanged with $u$ as
\begin{equation}
    \label{ExchangeF}
    Ex(u,S_i) = \{ v| (v \in S_i) \wedge (ch(v)=0) \wedge (0<-(sup(u)- sup(v))
    \leq sup_i)\}
\end{equation}
Here $sup_i$ gives the amount of available supply of node $S_i$. Using this definition we can present the correction procedure using the following pseudo-code:

 \begin{algorithmic}
\STATE
\REPEAT
\STATE Initialize  $N$ with all non-located nodes
\STATE $N_{out} = N$
  \FORALL{$u \in N$}
	  \FORALL{$S_i$ connected to $u$}
	  \STATE $Ex  = Ex(u,S_i)$ 	
	  \STATE
	  \IF{$(Ex \neq \emptyset)$ or $(sup(u)\leq sup_i)$}
	\STATE{$v = max\_dem(u,S_i)$}
	\STATE{$S_i = (S_i \setminus \{v \}) \cup \{u \}$ }
	\STATE{$N_{out} = (N_{out} \setminus \{u \}) \cup \{v \}$}
	\STATE{Update Auxiliary Structures}
	\STATE{Exit Subgraph loop}
    \ENDIF
   \ENDFOR
\ENDFOR
	\STATE $N = N_{out}$
\UNTIL{NoChange}
\end{algorithmic}

The main loop of the proposed correction procedure goes through all the non-located nodes $u$, and for each of them it checks if it can improve the amount of covered demand for one of the neighboring subgraphs. We consider a subgraph $S_i$ neighboring to node $u$ if there exists $v \in S_i$ such that $(u,v) \in E$. The correction is done greedily in the sense that the maximal level of improvement will be selected. More precisely, node $u$ will be exchanged with $v \in S_i$ giving the minimal value of $sup(v)$ with the constraint that the total demand will not be greater than the total available supply. Note that in some cases it may not be necessary to remove a node for $S_i$.

After each exchange it is necessary to update the auxiliary structures. Since it is possible that new improvements, become possible after the performed exchange operation, we also need to track the new unlocked nodes. In the pseudo code this is done using the set $N_{out}$. After all the non-located nodes are tested, we repeat the main loop for the new set of non-located nodes. The correction has been completed when no improvement can be made for any $u \in N$.

A specific case of the exchange correction is if we allow the exchange of a non-located node $u$ with a $v \in S_i$ in case $sup(u) = sup(v)$. It is obvious that we can not simply extend $Ex(u, S_i)$ by such $v$, since an endless cycle of first adding $u$ to $S_i$, and removing $v$ and in the following iteration we could just do the reverse if no further changes have been performed on $S_i$. On the other hand such an exchange is useful since it makes it possible to diversify the search. Because of this we will allow such exchanges, but they will be treated separately, in the sense that we will apply the algorithm given in the pseudo code, but exclusively with the change that $Ex(u, S_i)$ will only contain nodes $v$ such that $sup(v)=sup(u)$,  we will call this operation a switch non-located correction. The details of avoiding  cycles will be given in a later subsection.

\begin{figure*}[tcb]
\centering
\includegraphics[width=0.9\textwidth]{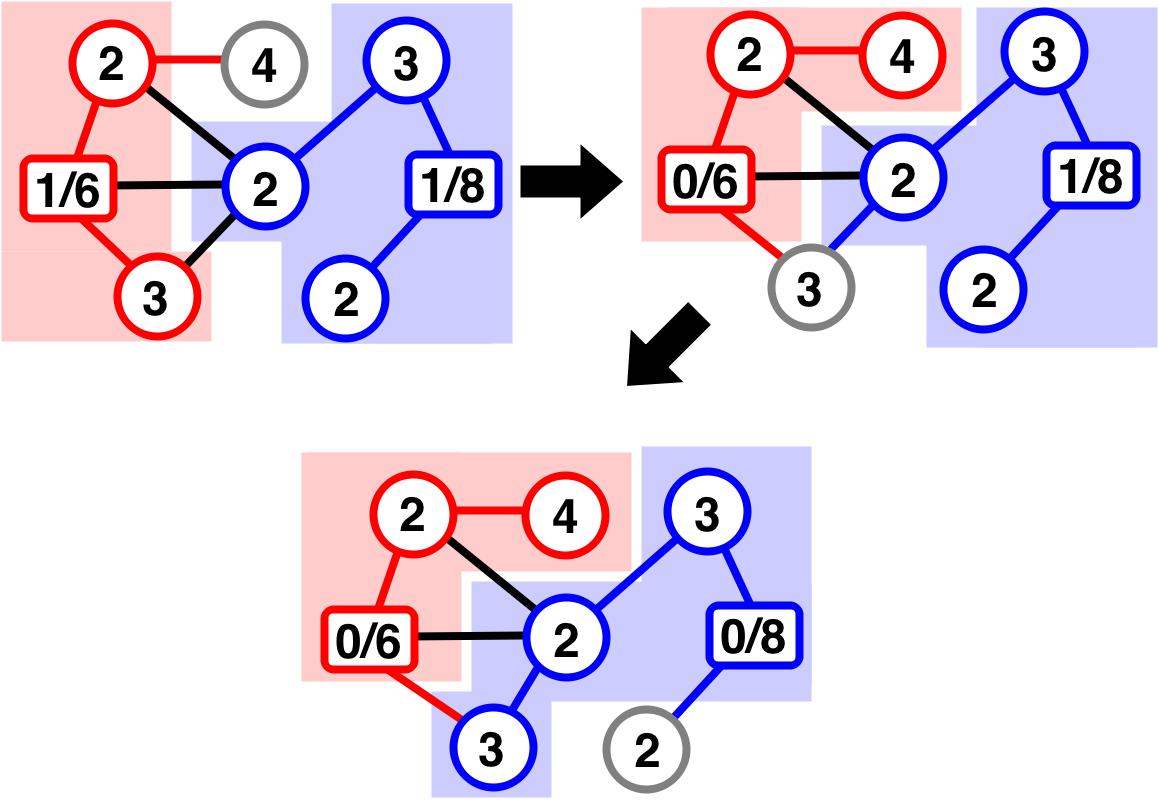}
\caption{Illustration of several steps of the non-located correction, and its ability to resolve complex exchanges between subgraphs when improving the solution.}
\label{fig:SimpleCorrection}
\end{figure*}

Although the procedure only performs tests in the relation of a free node to a single subgraph, by applying the consecutive corrections exchanges can occur between several subgraphs. By doing so this simple procedure can produce a significant level of improvement which we illustrate in Figure \ref{fig:SimpleCorrection}.


\subsubsection{Cutoff Correction}

\begin{figure*}[tcb]
\centering
\includegraphics[width=0.9\textwidth]{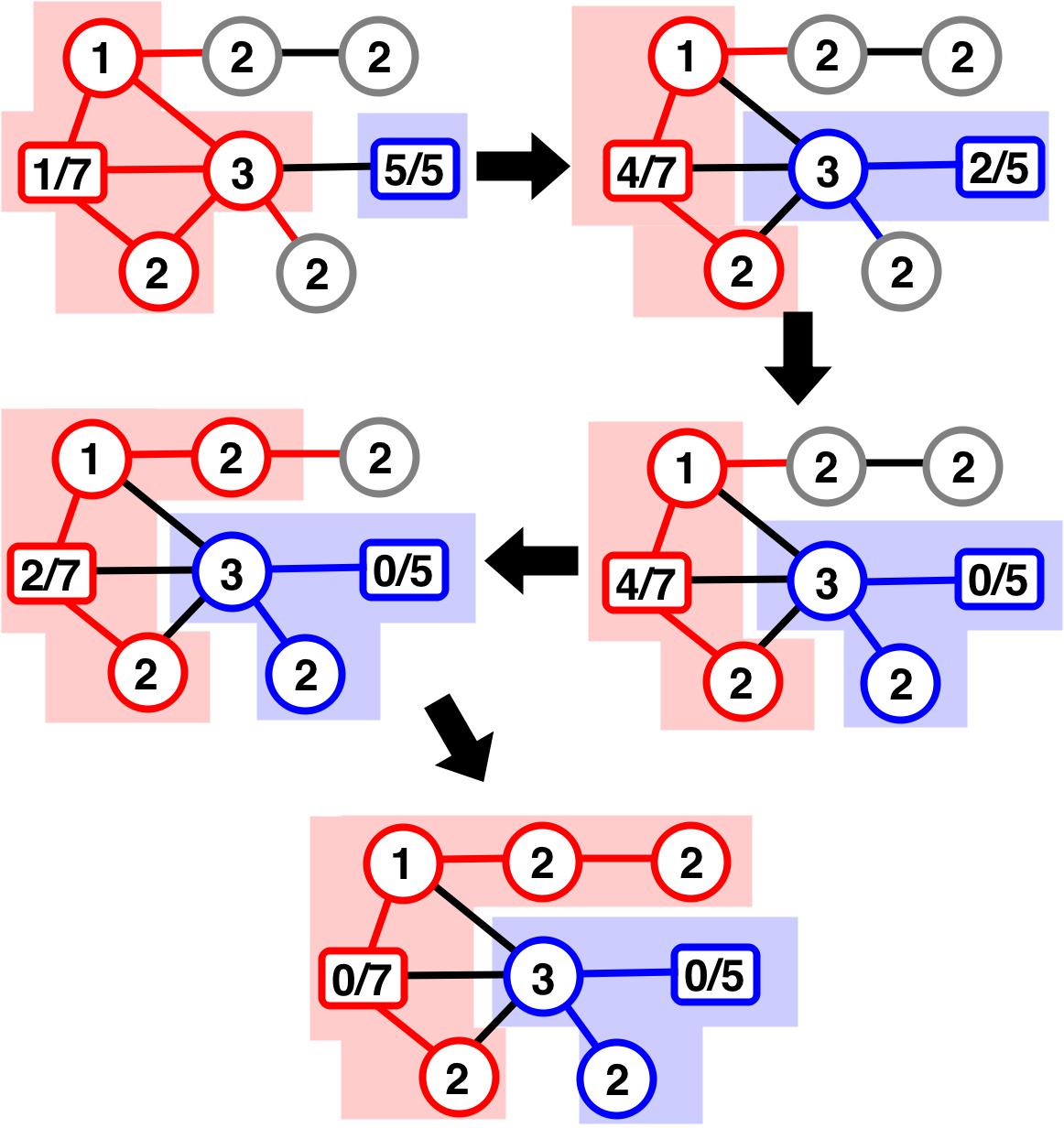}
\caption{Example of two consecutive applications of the cutoff correction. In this case the correction is first applied on the left subgraph, and the changes on the graph make it possible to improve the right subgraph in the second application of the correction.}
\label{fig:CutoffCorrection}
\end{figure*}

The focus of the second correction procedure is to improve a solution in which some supply nodes and their corresponding subgraphs have been cut from the rest of the graph at an early stage of the algorithm. One example of such a situation is given in Figure \ref{fig:CutoffCorrection}. In many cases it is not possible to improve such a partitioning using the previously presented correction procedure. The idea is to expand the subgraphs that have a high level of available supply by taking some demand nodes from neighboring subgraphs. This is done in the hope that  the expanded subgraph will get access to some non-located nodes or that the new corrected partitioning  can be improved using the previously presented correction procedure.

To formally present this correction procedure we first define the set of neighboring nodes to a subgraph
\begin{equation}
    \label{ExchangeF2nd}
    Nn(S_i) = \{ v|  (\exists u \in S_i)((v,u) \in E)\}
\end{equation}
Now we can define the set of nodes to which subgraph $S_i$ can be expanded in as follows:
\begin{equation}
    \label{ExchangeS}
    Ex(S_i) = \{ v|  (v \in Nn(S_i)) \wedge (ch(v)=0)  \wedge (sup(v)+ sup_i \ge 0) \}
\end{equation}
In Equation (\ref{ExchangeS}) the set $Ex(S_i)$ consists of all neighboring nodes that are leafs whose demand is not greater than the available supply $sup_i$ of subgraph $S_i$. With these definitions we can present the correction procedure using the following pseudocode.

 \begin{algorithmic}
\STATE
\STATE{$S = MaxSupplyExpandable(\Pi)$}
    \REPEAT
        \IF{($Ex(S) \neq \emptyset $)}
	    \STATE{$u = max\_dem(Ex(S))$}
	    \STATE{$S = S \cup \{u\}$}
	    \STATE{$Sub(u) = Sub(u) \setminus \{u\}$}
		\STATE{Update Auxiliary Structures}
		\ENDIF
    \UNTIL{NoChange}
\end{algorithmic}

As can be seen in the pseudo code we select the subgraph $S$ that has the maximal available supply and can be expanded or in other words $Ex(S) \neq \emptyset$. At each iteration of the  main loop we will expand $S$ with that node $u$ that has the highest demand, and we will always select a non-located node before one that is part of some subgraph. This procedure will be repeated until no more nodes can be add to $S$.

We illustrate this correction procedure in Figure \ref{fig:CutoffCorrection}, by showing the effect of two consecutive applications.

\subsubsection{Problem of Cycles}

One problem that can occur when using consecutive cutoff and switch corrections is that we can enter cycles. Here we use the term cycle for the situation when after $n$ corrections the partitioning is the same as before any corrections have been done. A simple example is when in the first correction some node $u \in S_i$ is moved to subgraph $S_j$, and in the following correction  $u$ is moved back to $S_i$. Note that such cycles can also include non-located nodes. It is evident that  cycles can be of an arbitrary length, but in our experience we have seen that they are very seldom longer than six correction operations.

In the implementation of the proposed algorithm we have used the following method for avoiding cycles. Let us assume the last $n$ corrections have resulted in a cycle. This means that a sequence of corrections  $C_1, C_2,..,C_n$ leaves the partitioning unchanged. A simple way of avoiding the repetition of the same cycle is not allowing correction $C_{n+1}$ to be equal to $C_1$. In practice this means we will remove the possibility of selecting  node $u$ that appears in correction $C_1$ when selecting $C_{n+1}$. Although this procedure escapes the majority of cycles, as we have seen in our tests, it can still get trapped in some. As a consequence, when implementing the proposed corrections, it is necessary to include a method for checking if the algorithm has started to stagnate. We will consider the algorithm stagnant if for $m$ corrections the covered demand has not increase.

For the sake of completeness we include a straightforward method to check if a cycle has happened in the last $n$ corrections.  For each node $u$ we know in which correction $C_i$ it first appears, we can store the pair $(S_j,u)$ if $u \in S_j$ before the correction is applied. In case that after applying $n$ corrections, we have  $u \in S_j$ for all the nodes that have been effected by some correction $C_i$, a cycle is completed. In our implementation of the proposed algorithm  we will be checking for all cycles up to length $n$.

\subsubsection{Combined Correction and Multi-Heuristic Approach}

From the definition of the cutoff correction it is evident that it does not necessarily  increase the amount of covered demand, but it changes the partitioning to one that can be improved using the the non-located node correction. To have the maximal effect of the proposed corrections it is necessary to have several consecutive applications of each. There are several potential orders in which these corrections can be applied to improve the quality of the solution acquired using the proposed greedy algorithm; through trial and error  we have found that best results are achieved when this is done as in the following pseudocode.

 \begin{algorithmic}
	\REPEAT
	\REPEAT
	  \IF{($CorrectionNonLocated()$)}
	     \STATE{$CorrectionSwitch()$}
	  \ENDIF
	\UNTIL{$(NoChangeNonLocated  \vee Stagnation)$}
	\REPEAT
	   \STATE{$CorrectionCutoff()$}
	\UNTIL{$(NoChangeCutOff  \vee Stagnation)$}
\UNTIL{($NoChange \vee Stagnation$)}
\end{algorithmic}

In the presented code functions \textit{CorrectionNonLocated()} and \textit{CorrectionCutoff()} perform the corrections as presented in the previous subsections, and return if any change has been performed on the graph. \textit{CorrectionSwitch()}  is used for the specific case of the non-located correction when only nodes with equal demand are allowed to be exchanged. We have found that the highest level of improvement is accomplished, when we  first exhaust all the non-located corrections, using the switch version of it to increase the potentially tested partitioning. In the second step we apply all the possible cutoff corrections. These two steps are repeated until no more changes to the partitioning are achieved.
A natural extension of the proposed algorithm is to use a varying number of nodes that are considered in the correction procedures. Similar approaches have proven very efficient on other graph problems.
As we have stated in the previous subsection, the method for avoiding cycles does not manage to detect all of them. Because of this it is necessary to include tests if the algorithm has started to stagnate in the sense that no increase to the level of covered demand has occurred  in the last $n$ iterations.

As mentioned in the introduction, when for a specific problem we have several competing heuristics it is a common practice to apply all of them and select the best found solution. This simple concept can effectively be applied in the case of  MPGSD, since we can combining  heuristics for different stages of the  algorithm. In this way we can explore a relatively small number of potentially good solutions. This approach becomes even more effective due to the use of the correction procedure, which in a sense behaves like a hill climbing method. Because of this the multiheuristic approach makes it possible to explore several local, potentially global, optimal solutions. In practice we have used three heuristics for the selection of the subgraph to be expanded, and four for the selection of nodes. In the case of node selection we have included an extra heuristic function which selects the node with the minimal demand. This  type of heuristic was not  disused in the corresponding subsection, due to the fact that it generally does not have a good performance. In that case of the multiheuristic approach used, since it effectively complements the other heuristics  and by combining it with the correction procedure, it can find good solutions that have been overseen by greedy algorithms based on the other heuristics.

\section{Tests and Results}
\label{MPGSDResults}

In this section we evaluate the performance of the proposed algorithm. We shall first give a comparison of the effect of using different heuristics for each stage of the algorithm. In our test we also examine the two proposed correction procedures and the multiheuristic approach.  The algorithm has been implemented in C\# using Microsoft Visual Studio 2012. The source code and the executive files are available at \url{http://mail.ipb.ac.rs/~rakaj/home/graphsd.htm}. The calculations have been done on a machine with Intel(R) Core(TM) i7-2630 QM CPU \@ 2.00 Ghz, 4GB of DDR3-1333 RAM, running on Microsoft Windows 7 Home Premium 64-bit.

With the goal of having a comprehensive evaluation of the proposed algorithm we have performed experiments on a wide range of graph sizes. We have conducted tests on graphs having  2-400 supply nodes and 6-8000 demand nodes. For each of the test sizes 40 different problem instances have been generated and we observe the average solution quality for each size. Such experiments have been performed on trees and general graphs. The goal of the conducted tests was to evaluate each heuristic,  correction methods and show the advantage of using the multiheuristic approach.

The problem instances (graphs) for $n$ supply and $m$ demand nodes have been generated using the following algorithm. First we generate an array containing $n+m$ integer random weights  uniformly distributed within the interval $[-10,-40]$. In case of general graphs, $(n+m)*2$ random edges would be added to the graph but making sure that the graph is connected. In case of the second type of graphs, i.e. trees, we would simply generate a random tree for $n+m$ nodes. For both types of graphs the next step was to select $n$ random nodes as seeds for $n$ subgraphs (partitions). The subgraphs are grown using an iterative method until $(n+m)*(0.95)$ nodes of the original graph are contained in one of the subgraphs. The growth of subgraph  $S_i$ has been performed by expanding it to a random neighboring node that does not belong to any of the other subgraphs. Finally, for each of the subgraphs $S_i$ a random vertex $v \in S_i$ is chosen and its weight $w$ is set using the following formula:
 \begin{equation}
    w = 	|\sum_{a \in S_i}sup(a)| +sup(v)
 \end{equation}

For each of the test graphs, generated using this method, the optimal solution is known and is equal to the sum of supplies of all supply nodes. Note that by using such a procedure it is possible to have partitionings that do not have $n$ subgraphs due to some supply nodes being cut off from the rest of the graph. We have excluded such graphs from our test cases. The generated test instances are available for download at \url{http://mail.ipb.ac.rs/~rakaj/home/graphsd.htm}.

In the first group of tests a comparison of the effect of using different heuristics for selecting subgraph $S_i$ is observed. The results of this type of experiments are given in Tables \ref{table:GenSelSub}, \ref{table:TreeSelSub}  for general graphs and trees. For each of the heuristics the same heuristic $hn1$, node with highest demand, was used for node selection. In these tables we observe the average normalized error of the found solution compared to the optimal one, which is known due to the method of generating problem instances. More precisely, for each of the 40 test instances, for one graph size, we calculate the normalized error in percent $(Optimal - found)/Optimal *100$, and give the average value of such errors in Tables \ref{table:GenSelSub}, \ref{table:TreeSelSub}. To have a better analysis of the proposed heuristics we also observe the standard deviation, and maximal error.

\begin{table*}[htb]
\footnotesize
\center
\caption{\label{table:GenSelSub}Comparison of heuristics for subgraph selection for general graphs }
\begin{tabularx}{\linewidth}{X*{10}{c}}
\toprule
Sup X Dem&    \multicolumn{3}{c}{$hs_1$}  & \multicolumn{3}{c}{$hs_2$}& \multicolumn{3}{c}{$hs_3$} \\

          &    Avg    &   Max & StDev &     Avg    &    Max& StDev &     Avg    &    Max& StDev \\
\midrule
2 X 6 & 7.4 & 46.1 & 8.7 & \underline{3.9} & \underline{19.3} & 5.6 & 5.6 & \underline{19.3} & 6.1 \\
2 X 10 & 5.6 & 23.1 & 4.4 & \underline{4.3} & \underline{11.8} & 2.6 & 4.5 & \underline{11.8} & 3.1 \\
2 X 20 & 1.8 & 4.2 & 1.1 & \underline{1.7} & \underline{4.1} & 1.1 & 2.1 & 5.2 & 1.3 \\
2 X 40 & 0.8 & 1.7 & 0.5 & 0.7 & 2.1 & 0.5 & \underline{0.6} & \underline{1.7} & 0.4 \\
\midrule
5 X 15 & 10.9 & 38.8 & 7.8 & \underline{6.5} & \underline{20.0} & 4.8 & 7.9 & 20.1 & 5.3 \\
5 X 25 & 7.9 & 34.6 & 6.0 & \underline{4.5} & 15.6 & 3.3 & 4.9 & \underline{13.1} & 2.5 \\
5 X 50 & 3.9 & 10.3 & 2.6 & \underline{1.9} & \underline{5.2} & 1.0 & 2.6 & 8.1 & 1.3 \\
5 X 100 & 2.0 & 13.6 & 2.5 & 0.9 & 4.8 & 0.8 & \underline{0.9} & \underline{3.7} & 0.6 \\
\midrule
10 X 30 & 11.5 & 23.9 & 4.4 & \underline{6.7} & \underline{15.2} & 3.4 & 8.4 & 18.1 & 3.5 \\
10 X 50 & 7.4 & 14.2 & 2.8 & 5.2 & 15.0 & 3.0 & \underline{5.0} & \underline{12.2} & 2.1 \\
10 X 100 & 3.9 & 13.1 & 2.4 & 2.3 & 9.5 & 1.6 & \underline{2.3} & \underline{5.2} & 0.9 \\
10 X 200 & 2.5 & 13.0 & 2.8 & 1.5 & 10.6 & 1.9 & \underline{1.1} & \underline{5.9} & 0.9 \\
\midrule
25 X 75 & 12.1 & 19.2 & 3.2 & \underline{8.2} & 16.8 & 2.9 & 9.1 & \underline{14.6} & 2.9 \\
25 X 125 & 8.6 & 13.6 & 2.1 & \underline{5.5} & \underline{8.9} & 1.6 & 6.0 & 10.3 & 1.6 \\
25 X 250 & 4.6 & 8.7 & 1.5 & \underline{2.8} & 7.7 & 1.4 & 2.9 & \underline{5.5} & 0.9 \\
25 X 500 & 2.8 & 6.1 & 1.4 & 1.5 & 4.9 & 1.1 & \underline{1.3} & \underline{3.2} & 0.6 \\
\midrule
50 X 150 & 12.0 & 15.6 & 1.9 & \underline{8.0} & \underline{12.2} & 2.1 & 8.6 & 12.2 & 1.8 \\
50 X 250 & 8.8 & 10.8 & 1.3 & \underline{5.9} & 10.4 & 1.6 & 6.1 & \underline{10.1} & 1.3 \\
50 X 500 & 4.6 & 7.4 & 1.3 & \underline{2.6} & 4.5 & 0.9 & 2.6 & \underline{3.7} & 0.6 \\
50 X 1000 & 3.1 & 6.0 & 1.0 & \underline{1.2} & 2.9 & 0.6 & 1.3 & \underline{2.8} & 0.5 \\
\midrule
100 X 300 & 11.7 & 14.6 & 1.5 & \underline{7.9} & \underline{10.6} & 1.3 & 8.7 & 11.0 & 1.1 \\
100 X 500 & 8.8 & 11.6 & 1.1 & \underline{5.7} & 9.0 & 1.1 & 6.0 & \underline{8.2} & 0.9 \\
100 X 1000 & 4.7 & 7.0 & 0.9 & \underline{2.9} & 5.6 & 0.8 & 2.9 & \underline{4.3} & 0.5 \\
100 X 2000 & 3.0 & 4.6 & 0.7 & 1.5 & 3.0 & 0.6 & \underline{1.4} & \underline{2.3} & 0.4 \\
\midrule
200 X 600 & 12.1 & 14.3 & 1.0 & \underline{8.0} & \underline{9.7} & 0.9 & 8.9 & 11.8 & 0.9 \\
200 X 1000 & 8.8 & 10.8 & 0.7 & \underline{5.8} & 7.4 & 0.6 & 6.2 & \underline{7.2} & 0.5 \\
200 X 2000 & 4.9 & 6.2 & 0.5 & \underline{2.9} & 4.0 & 0.5 & 2.9 & \underline{3.8} & 0.3 \\
200 X 4000 & 3.1 & 4.2 & 0.4 & 1.5 & 2.5 & 0.4 & \underline{1.4} & \underline{2.2} & 0.3 \\
\midrule
400 X 1200 & 11.8 & 13.4 & 0.6 & \underline{7.9} & \underline{9.2} & 0.6 & 8.7 & 9.8 & 0.5 \\
400 X 2000 & 8.7 & 10.0 & 0.5 & \underline{5.9} & 7.0 & 0.6 & 6.2 & \underline{6.8} & 0.3 \\
400 X 4000 & 4.8 & 5.8 & 0.5 & \underline{2.8} & \underline{3.4} & 0.3 & 2.9 & 3.7 & 0.3 \\
400 X 8000 & 3.0 & 3.7 & 0.3 & 1.5 & 2.0 & 0.3 & \underline{1.4} & \underline{1.8} & 0.2 \\
\bottomrule
\end{tabularx}
\end{table*}

\begin{table*}[htb]
\footnotesize
\center
\caption{\label{table:TreeSelSub}Comparison of heuristics for subgraph selection for trees }
\begin{tabularx}{\linewidth}{X*{10}{c}}
\toprule
Sup X Dem&    \multicolumn{3}{c}{$hs_1$}  & \multicolumn{3}{c}{$hs_2$}& \multicolumn{3}{c}{$hs_3$} \\

          &    Avg    &   Max & StDev &     Avg    &    Max& StDev &     Avg    &    Max& StDev \\
\midrule
2 X 6 & 1.7 & 26.4 & 5.9 & \underline{0.0} & \underline{0.0} & 0.0 & 0.5 & 18.2 & 2.8 \\
2 X 10 & 5.5 & 35.1 & 8.0 & 3.5 & 51.1 & 8.2 & \underline{3.2} & \underline{20.5} & 5.1 \\
2 X 20 & 8.7 & \underline{28.9} & 9.1 & 5.1 & 40.6 & 9.7 & \underline{5.0} & 29.8 & 7.0 \\
2 X 40 & 6.1 & 30.6 & 7.7 & 3.5 & 41.1 & 8.9 & \underline{2.7} & \underline{11.7} & 3.5 \\
\midrule
5 X 15 & 8.5 & 27.2 & 8.7 & 3.8 & \underline{23.4} & 6.3 & \underline{2.4} & \underline{23.4} & 4.8 \\
5 X 25 & 7.9 & \underline{21.8} & 6.0 & 7.7 & 47.6 & 11.1 & \underline{5.2} & 25.5 & 5.5 \\
5 X 50 & 10.6 & 29.6 & 7.0 & 8.3 & 41.4 & 10.1 & \underline{5.3} & \underline{15.8} & 4.5 \\
5 X 100 & 16.4 & 50.9 & 11.2 & 15.8 & 43.7 & 13.1 & \underline{6.2} & \underline{17.1} & 4.4 \\
10 X 30 & 8.7 & 27.2 & 6.4 & 4.3 & 19.2 & 5.5 & \underline{4.2} & \underline{17.2} & 4.6 \\
\midrule
10 X 50 & 9.7 & 29.5 & 5.6 & 6.1 & 24.6 & 5.7 & \underline{4.4} & \underline{11.6} & 3.2 \\
10 X 100 & 11.4 & 26.7 & 6.3 & 9.6 & 30.1 & 7.0 & \underline{5.7} & \underline{14.9} & 3.8 \\
10 X 200 & 13.9 & 26.0 & 6.6 & 13.1 & 32.6 & 8.1 & \underline{7.6} & \underline{21.6} & 4.7 \\
\midrule
25 X 75 & 9.5 & 22.5 & 4.9 & \underline{3.6} & \underline{12.9} & 3.4 & 5.1 & 13.6 & 3.4 \\
25 X 125 & 10.8 & 17.3 & 3.8 & 6.8 & 17.8 & 4.2 & \underline{5.3} & \underline{11.4} & 2.2 \\
25 X 250 & 10.7 & 20.2 & 3.2 & 8.6 & 19.6 & 4.0 & \underline{5.7} & \underline{11.5} & 2.1 \\
25 X 500 & 11.6 & 19.4 & 3.9 & 11.1 & 24.1 & 4.3 & \underline{6.1} & \underline{12.3} & 2.5 \\
50 X 150 & 8.7 & 17.0 & 2.9 & 4.4 & 8.9 & 2.2 & \underline{4.2} & \underline{8.9} & 1.9 \\
\midrule
50 X 250 & 10.2 & 18.7 & 3.1 & 6.2 & 12.3 & 2.5 & \underline{5.7} & \underline{12.2} & 2.0 \\
50 X 500 & 11.9 & 18.8 & 3.0 & 8.5 & 13.1 & 2.3 & \underline{6.9} & \underline{11.9} & 2.0 \\
50 X 1000 & 12.8 & 18.3 & 2.4 & 11.3 & 16.9 & 3.1 & \underline{7.1} & \underline{10.8} & 1.5 \\
\midrule
100 X 300 & 9.8 & 14.1 & 2.0 & \underline{4.8} & 8.8 & 1.8 & 5.3 & \underline{8.3} & 1.3 \\
100 X 500 & 10.3 & 14.4 & 1.8 & 6.1 & 10.8 & 1.8 & \underline{5.7} & \underline{8.9} & 1.3 \\
100 X 1000 & 11.2 & 14.5 & 1.8 & 8.4 & 14.0 & 2.1 & \underline{6.5} & \underline{8.2} & 1.2 \\
100 X 2000 & 12.1 & 17.5 & 1.9 & 10.3 & 13.5 & 2.0 & \underline{6.8} & \underline{9.8} & 1.3 \\
\midrule
200 X 600 & 9.5 & 13.7 & 1.5 & \underline{4.5} & 8.2 & 1.6 & 5.2 & \underline{7.7} & 1.2 \\
200 X 1000 & 10.2 & 12.7 & 1.2 & 6.2 & \underline{7.9} & 1.0 & \underline{5.9} & 8.0 & 0.9 \\
200 X 2000 & 11.4 & 14.8 & 1.5 & 8.7 & 11.9 & 1.3 & \underline{6.6} & \underline{8.4} & 0.9 \\
200 X 4000 & 12.0 & 15.1 & 1.5 & 10.2 & 13.5 & 1.5 & \underline{7.0} & \underline{9.3} & 0.9 \\
\midrule
400 X 1200 & 9.3 & 11.8 & 0.9 & \underline{4.6} & \underline{6.7} & 0.7 & 5.4 & 6.9 & 0.7 \\
400 X 2000 & 10.2 & 12.6 & 1.0 & 6.3 & 8.4 & 0.9 & \underline{6.0} & \underline{7.3} & 0.7 \\
400 X 4000 & 11.2 & 13.1 & 1.0 & 8.7 & 11.5 & 0.9 & \underline{6.9} & \underline{8.2} & 0.6 \\
400 X 8000 & 11.8 & 13.6 & 1.0 & 10.2 & 12.8 & 0.9 & \underline{7.1} & \underline{8.7} & 0.6 \\
\bottomrule
\end{tabularx}
\end{table*}

\begin{table*}[htb]
\footnotesize
\center
\caption{\label{table:GenSelNode}Comparison of heuristics for node selection for general graphs}
\begin{tabularx}{\linewidth}{X*{10}{c}}
\toprule
Sup X Dem&    \multicolumn{3}{c}{$hs_1$}  & \multicolumn{3}{c}{$hs_2$}& \multicolumn{3}{c}{$hs_3$} \\

          &    Avg    &   Max & StDev &     Avg    &    Max& StDev &     Avg    &    Max& StDev \\
\midrule
2 X 6 & \underline{3.9} & \underline{19.3} & 5.6 & 10.8 & 31.6 & 9.7 & 4.0 & \underline{19.3} & 5.3 \\
2 X 10 & \underline{4.3} & \underline{11.8} & 2.6 & 8.1 & 18.1 & 4.3 & 4.9 & 12.3 & 2.5 \\
2 X 20 & \underline{1.7} & \underline{4.1} & 1.1 & 4.2 & 14.7 & 2.7 & 1.7 & 5.2 & 1.3 \\
2 X 40 & \underline{0.7} & 2.1 & 0.5 & 1.7 & 4.0 & 1.0 & 0.8 & \underline{2.0} & 0.5 \\
\midrule
5 X 15 & \underline{6.5} & \underline{20.0} & 4.8 & 11.9 & 28.3 & 5.7 & 8.7 & 26.4 & 6.0 \\
5 X 25 & \underline{4.5} & \underline{15.6} & 3.3 & 7.7 & 17.7 & 3.4 & 5.5 & 16.4 & 3.4 \\
5 X 50 & \underline{1.9} & \underline{5.2} & 1.0 & 3.6 & 9.9 & 1.8 & 3.1 & 10.5 & 2.2 \\
5 X 100 & \underline{0.9} & \underline{4.8} & 0.8 & 2.4 & 7.5 & 1.5 & 1.8 & 14.4 & 2.7 \\
\midrule
10 X 30 & \underline{6.7} & \underline{15.2} & 3.4 & 10.5 & 19.3 & 3.4 & 7.1 & 17.4 & 3.3 \\
10 X 50 & \underline{5.2} & 15.0 & 3.0 & 7.9 & 18.6 & 3.2 & 6.3 & \underline{13.1} & 2.5 \\
10 X 100 & \underline{2.3} & \underline{9.5} & 1.6 & 4.5 & 9.6 & 2.1 & 3.2 & 11.3 & 2.1 \\
10 X 200 & \underline{1.5} & 10.6 & 1.9 & 2.8 & 8.9 & 1.6 & 2.3 & \underline{8.0} & 1.8 \\
\midrule
25 X 75 & \underline{8.2} & \underline{16.8} & 2.9 & 11.2 & 17.0 & 2.6 & 8.8 & 19.5 & 3.4 \\
25 X 125 & \underline{5.5} & \underline{8.9} & 1.6 & 8.5 & 17.6 & 2.2 & 6.2 & 10.4 & 1.7 \\
25 X 250 & \underline{2.8} & 7.7 & 1.4 & 4.7 & 7.0 & 1.1 & 3.6 & \underline{6.9} & 1.5 \\
25 X 500 & \underline{1.5} & \underline{4.9} & 1.1 & 3.0 & 6.4 & 1.2 & 2.2 & 5.0 & 1.0 \\
\midrule
50 X 150 & \underline{8.0} & \underline{12.2} & 2.1 & 11.2 & 15.5 & 2.2 & 8.9 & 15.3 & 2.8 \\
50 X 250 & \underline{5.9} & \underline{10.4} & 1.6 & 9.0 & 12.3 & 1.6 & 6.7 & 10.4 & 1.5 \\
50 X 500 & \underline{2.6} & \underline{4.5} & 0.9 & 5.0 & 7.0 & 1.1 & 3.4 & 5.7 & 0.9 \\
50 X 1000 & \underline{1.2} & \underline{2.9} & 0.6 & 3.0 & 4.4 & 0.7 & 2.3 & 3.4 & 0.6 \\
\midrule
100 X 300 & \underline{7.9} & \underline{10.6} & 1.3 & 10.7 & 13.5 & 1.4 & 8.7 & 11.9 & 1.6 \\
100 X 500 & \underline{5.7} & \underline{9.0} & 1.1 & 8.6 & 10.2 & 0.9 & 6.6 & 9.2 & 1.2 \\
100 X 1000 & \underline{2.9} & 5.6 & 0.8 & 5.1 & 6.6 & 0.7 & 3.7 & \underline{5.6} & 0.8 \\
100 X 2000 & \underline{1.5} & \underline{3.0} & 0.6 & 3.3 & 4.8 & 0.7 & 2.4 & 4.1 & 0.7 \\
\midrule
200 X 600 & \underline{8.0} & \underline{9.7} & 0.9 & 11.2 & 13.6 & 1.0 & 8.6 & 11.5 & 1.0 \\
200 X 1000 & \underline{5.8} & \underline{7.4} & 0.6 & 8.6 & 9.8 & 0.6 & 6.7 & 8.6 & 0.9 \\
200 X 2000 & \underline{2.9} & \underline{4.0} & 0.5 & 5.0 & 6.5 & 0.6 & 3.7 & 5.1 & 0.6 \\
200 X 4000 & \underline{1.5} & \underline{2.5} & 0.4 & 3.2 & 4.0 & 0.4 & 2.3 & 3.6 & 0.5 \\
\midrule
400 X 1200 & \underline{7.9} & \underline{9.2} & 0.6 & 11.3 & 12.9 & 0.7 & 8.7 & 10.3 & 0.8 \\
400 X 2000 & \underline{5.9} & \underline{7.0} & 0.6 & 8.7 & 10.4 & 0.6 & 6.6 & 7.8 & 0.5 \\
400 X 4000 & \underline{2.8} & \underline{3.4} & 0.3 & 5.1 & 6.0 & 0.4 & 3.7 & 4.9 & 0.4 \\
400 X 8000 & \underline{1.5} & \underline{2.0} & 0.3 & 3.2 & 3.9 & 0.3 & 2.4 & 3.0 & 0.3 \\

\bottomrule
\end{tabularx}
\end{table*}

\begin{table*}[htb]
\footnotesize
\center
\caption{\label{table:TreeSelNode}Comparison of heuristics for node selection for trees}
\begin{tabularx}{\linewidth}{X*{10}{c}}
\toprule
Sup X Dem&    \multicolumn{3}{c}{$hs_1$}  & \multicolumn{3}{c}{$hs_2$}& \multicolumn{3}{c}{$hs_3$} \\

          &    Avg    &   Max & StDev &     Avg    &    Max& StDev &     Avg    &    Max& StDev \\
\midrule
2 X 6 & 0.5 & 18.2 & 2.8 & 2.8 & 24.5 & 7.5 & \underline{0.0} & \underline{0.0} & 0.0 \\
2 X 10 & 3.2 & 20.5 & 5.1 & 2.1 & \underline{9.9} & 3.4 & \underline{1.8} & 20.5 & 3.9 \\
2 X 20 & 5.0 & 29.8 & 7.0 & \underline{3.5} & \underline{26.9} & 5.8 & 4.2 & 28.6 & 7.3 \\
2 X 40 & 2.7 & \underline{11.7} & 3.5 & 3.2 & 21.8 & 4.9 & \underline{2.6} & 21.2 & 4.8 \\
\midrule
5 X 15 & \underline{2.4} & \underline{23.4} & 4.8 & 5.4 & 30.4 & 7.1 & 3.2 & 28.1 & 6.5 \\
5 X 25 & 5.2 & \underline{25.5} & 5.5 & 6.3 & \underline{25.5} & 6.2 & \underline{4.4} & \underline{25.5} & 5.4 \\
5 X 50 & 5.3 & 15.8 & 4.5 & 5.2 & 16.9 & 4.9 & \underline{4.1} & \underline{14.2} & 3.9 \\
5 X 100 & 6.2 & 17.1 & 4.4 & 5.6 & 20.1 & 5.1 & \underline{5.1} & \underline{16.4} & 4.8 \\
\midrule
10 X 30 & 4.2 & 17.2 & 4.6 & 4.7 & 18.0 & 4.8 & \underline{3.2} & \underline{16.0} & 3.8 \\
10 X 50 & 4.4 & 11.6 & 3.2 & 4.0 & 11.2 & 3.0 & \underline{3.0} & \underline{8.7} & 2.6 \\
10 X 100 & 5.7 & 14.9 & 3.8 & 4.9 & 14.1 & 3.3 & \underline{3.9} & \underline{11.7} & 3.1 \\
10 X 200 & 7.6 & 21.6 & 4.7 & 6.3 & 17.2 & 3.6 & \underline{5.5} & \underline{12.2} & 3.3 \\
\midrule
25 X 75 & 5.1 & 13.6 & 3.4 & 5.3 & 16.7 & 3.1 & \underline{3.1} & \underline{11.0} & 2.7 \\
25 X 125 & 5.3 & 11.4 & 2.2 & 4.8 & 11.1 & 2.6 & \underline{4.0} & \underline{10.9} & 2.6 \\
25 X 250 & 5.7 & 11.5 & 2.1 & 4.9 & \underline{8.5} & 1.8 & \underline{4.4} & 9.9 & 2.1 \\
25 X 500 & 6.1 & 12.3 & 2.5 & 5.9 & 13.9 & 2.4 & \underline{5.6} & \underline{10.2} & 2.0 \\
\midrule
50 X 150 & 4.2 & 8.9 & 1.9 & 5.7 & 11.9 & 2.1 & \underline{3.1} & \underline{7.3} & 1.8 \\
50 X 250 & 5.7 & 12.2 & 2.0 & 5.9 & 9.3 & 1.6 & \underline{4.4} & \underline{9.2} & 1.8 \\
50 X 500 & 6.9 & 11.9 & 2.0 & 6.0 & \underline{9.3} & 1.7 & \underline{5.3} & 12.0 & 2.1 \\
50 X 1000 & 7.1 & 10.8 & 1.5 & 6.1 & \underline{9.5} & 1.7 & \underline{6.1} & 9.6 & 1.8 \\
\midrule
100 X 300 & 5.3 & 8.3 & 1.3 & 6.1 & 10.2 & 1.9 & \underline{3.8} & \underline{7.7} & 1.4 \\
100 X 500 & 5.7 & 8.9 & 1.3 & 5.8 & 8.4 & 1.2 & \underline{4.4} & \underline{6.9} & 1.2 \\
100 X 1000 & 6.5 & \underline{8.2} & 1.2 & 6.0 & 8.8 & 1.4 & \underline{5.5} & 9.8 & 1.5 \\
100 X 2000 & 6.8 & 9.8 & 1.3 & \underline{6.3} & \underline{8.3} & 1.2 & 6.4 & 10.2 & 1.5 \\
\midrule
200 X 600 & 5.2 & 7.7 & 1.2 & 5.4 & 7.2 & 1.0 & \underline{3.4} & \underline{4.9} & 0.9 \\
200 X 1000 & 5.9 & 8.0 & 0.9 & 5.7 & 7.7 & 0.9 & \underline{4.4} & \underline{7.3} & 1.0 \\
200 X 2000 & 6.6 & 8.4 & 0.9 & 6.1 & 8.6 & 1.0 & \underline{5.5} & \underline{8.1} & 1.0 \\
200 X 4000 & 7.0 & 9.3 & 0.9 & 6.4 & \underline{8.7} & 1.0 & \underline{6.3} & 9.5 & 1.2 \\
\midrule
400 X 1200 & 5.4 & 6.9 & 0.7 & 6.0 & 7.6 & 0.7 & \underline{3.8} & \underline{4.8} & 0.5 \\
400 X 2000 & 6.0 & 7.3 & 0.7 & 5.8 & 7.3 & 0.6 & \underline{4.5} & \underline{5.8} & 0.6 \\
400 X 4000 & 6.9 & 8.2 & 0.6 & 6.0 & 7.1 & 0.5 & \underline{5.6} & \underline{6.4} & 0.5 \\
400 X 8000 & 7.1 & 8.7 & 0.6 & \underline{6.4} & \underline{7.6} & 0.7 & 6.6 & 8.2 & 0.5 \\
\bottomrule
\end{tabularx}
\end{table*}

\begin{table*}[htb]
\footnotesize
\center
\caption{\label{table:GenCor}Comparison of correction heuristics for node selection for general graphs}
\begin{tabularx}{385pt}{X*{13}{c}}

\toprule
Sup X Dem&  \multicolumn{4}{c}{$Error$} & \multicolumn{4}{c}{$Max$}  & \multicolumn{4}{c}{$Time$} \\

         & Gr &NL &Com & Mult& Gr &NL &Com & Mult& Gr &NL &Com & Mult \\
\midrule
2 X 6 & 3.9 & 3.9 & 3.9 & 0.5 & 19.3 & 19.3 & 19.3 & 10.2 & 0.0 & 0.0 & 0.0 & 0.0 \\
2 X 10 & 4.3 & 4.2 & 4.2 & 0.8 & 11.8 & 11.8 & 11.8 & 6.0 & 0.0 & 0.0 & 0.0 & 0.0 \\
2 X 20 & 1.7 & 1.5 & 1.5 & 0.1 & 4.1 & 3.7 & 3.7 & 0.8 & 0.0 & 0.0 & 0.0 & 0.0 \\
2 X 40 & 0.7 & 0.5 & 0.5 & 0.0 & 2.1 & 2.1 & 2.1 & 0.0 & 0.0 & 0.0 & 0.0 & 0.0 \\
\midrule
5 X 15 & 6.5 & 5.9 & 5.2 & 2.0 & 20.0 & 20.0 & 14.9 & 9.1 & 0.0 & 0.0 & 0.0 & 0.0 \\
5 X 25 & 4.5 & 3.4 & 3.2 & 1.3 & 15.6 & 8.1 & 8.1 & 3.4 & 0.0 & 0.0 & 0.0 & 0.0 \\
5 X 50 & 1.9 & 1.5 & 1.4 & 0.4 & 5.2 & 2.8 & 2.8 & 1.2 & 0.0 & 0.0 & 0.0 & 0.0 \\
5 X 100 & 0.9 & 0.7 & 0.5 & 0.0 & 4.8 & 2.8 & 1.2 & 0.1 & 0.0 & 0.0 & 0.0 & 0.2 \\
\midrule
10 X 30 & 6.7 & 5.6 & 4.7 & 2.0 & 15.2 & 11.5 & 11.0 & 6.3 & 0.0 & 0.0 & 0.0 & 0.0 \\
10 X 50 & 5.2 & 4.5 & 3.6 & 1.9 & 15.0 & 12.2 & 7.3 & 4.3 & 0.0 & 0.0 & 0.0 & 0.0 \\
10 X 100 & 2.3 & 1.9 & 1.4 & 0.6 & 9.5 & 9.5 & 2.8 & 1.0 & 0.0 & 0.0 & 0.0 & 0.2 \\
10 X 200 & 1.5 & 1.2 & 0.6 & 0.1 & 10.6 & 10.1 & 2.8 & 0.2 & 0.0 & 0.0 & 0.1 & 0.9 \\
\midrule
25 X 75 & 8.2 & 7.4 & 6.0 & 3.6 & 16.8 & 16.8 & 10.5 & 6.8 & 0.0 & 0.0 & 0.0 & 0.2 \\
25 X 125 & 5.5 & 4.8 & 3.9 & 2.6 & 8.9 & 8.9 & 6.3 & 3.7 & 0.0 & 0.0 & 0.0 & 0.5 \\
25 X 250 & 2.8 & 2.4 & 1.6 & 0.9 & 7.7 & 7.4 & 4.3 & 1.5 & 0.0 & 0.0 & 0.1 & 1.4 \\
25 X 500 & 1.5 & 1.3 & 0.7 & 0.2 & 4.9 & 4.9 & 2.5 & 0.3 & 0.0 & 0.1 & 0.2 & 3.5 \\
\midrule
50 X 150 & 8.0 & 6.8 & 5.5 & 4.0 & 12.2 & 11.1 & 8.6 & 5.8 & 0.0 & 0.0 & 0.1 & 0.9 \\
50 X 250 & 5.9 & 5.2 & 4.2 & 3.2 & 10.4 & 9.6 & 8.8 & 4.2 & 0.0 & 0.0 & 0.2 & 1.9 \\
50 X 500 & 2.6 & 2.3 & 1.6 & 1.0 & 4.5 & 4.1 & 3.0 & 1.5 & 0.1 & 0.1 & 0.3 & 4.2 \\
50 X 1000 & 1.2 & 1.1 & 0.7 & 0.2 & 2.9 & 2.7 & 1.9 & 0.5 & 0.2 & 0.3 & 0.6 & 10.1 \\
\midrule
100 X 300 & 7.9 & 7.1 & 6.0 & 4.8 & 10.6 & 8.9 & 7.9 & 7.0 & 0.1 & 0.1 & 0.3 & 3.0 \\
100 X 500 & 5.7 & 5.1 & 4.2 & 3.5 & 9.0 & 8.4 & 6.4 & 5.1 & 0.1 & 0.2 & 0.4 & 5.5 \\
100 X 1000 & 2.9 & 2.5 & 1.8 & 1.2 & 5.6 & 4.8 & 3.0 & 1.6 & 0.3 & 0.4 & 0.9 & 12.5 \\
100 X 2000 & 1.5 & 1.3 & 0.7 & 0.3 & 3.0 & 2.8 & 1.5 & 0.8 & 0.7 & 0.9 & 1.9 & 30.4 \\
\midrule
200 X 600 & 8.0 & 7.1 & 6.4 & 5.3 & 9.7 & 8.6 & 8.6 & 6.9 & 0.3 & 0.4 & 0.7 & 9.4 \\
200 X 1000 & 5.8 & 5.2 & 4.5 & 3.9 & 7.4 & 6.4 & 6.2 & 4.4 & 0.6 & 0.7 & 1.3 & 19.6 \\
200 X 2000 & 2.9 & 2.5 & 2.0 & 1.3 & 4.0 & 3.6 & 2.8 & 1.7 & 1.4 & 1.6 & 3.4 & 48.3 \\
200 X 4000 & 1.5 & 1.3 & 0.8 & 0.4 & 2.5 & 2.2 & 1.4 & 0.8 & 2.6 & 3.3 & 6.3 & 98.7 \\
\midrule
400 X 1200 & 7.9 & 7.1 & 6.4 & 5.6 & 9.2 & 8.5 & 8.2 & 6.8 & 1.4 & 1.7 & 2.5 & 33.8 \\
400 X 2000 & 5.9 & 5.3 & 4.6 & 4.0 & 7.0 & 6.3 & 5.5 & 4.5 & 2.3 & 2.9 & 4.5 & 61.2 \\
400 X 4000 & 2.8 & 2.5 & 2.0 & 1.4 & 3.4 & 3.1 & 2.7 & 1.9 & 4.8 & 5.9 & 9.4 & 145.7 \\
400 X 8000 & 1.5 & 1.3 & 0.8 & 0.4 & 2.0 & 1.8 & 1.2 & 0.7 & 9.8 & 13.4 & 22.0 & 348.5 \\
\bottomrule
\end{tabularx}
\end{table*}

\begin{table*}[htb]
\footnotesize
\center
\caption{\label{table:TreeCor}Comparison of corrections for trees }
\begin{tabularx}{385pt}{X*{13}{c}}

\toprule
Sup X Dem&  \multicolumn{4}{c}{$Error$} & \multicolumn{4}{c}{$Max$}  & \multicolumn{4}{c}{$Time$} \\

         & Gr &NL &Com & Mult& Gr &NL &Com & Mult& Gr &NL &Com & Mult \\
\midrule
2 X 6 & 0.0 & 0.0 & 0.0 & 0.0 & 0.0 & 0.0 & 0.0 & 0.0 & 0.0 & 0.0 & 0.0 & 0.0 \\
2 X 10 & 1.8 & 1.6 & 1.1 & 0.1 & 20.5 & 20.5 & 8.8 & 3.4 & 0.0 & 0.0 & 0.0 & 0.0 \\
2 X 20 & 4.2 & 3.9 & 1.8 & 0.1 & 28.6 & 27.5 & 27.5 & 3.1 & 0.0 & 0.0 & 0.0 & 0.0 \\
2 X 40 & 2.6 & 2.4 & 2.4 & 0.3 & 21.2 & 21.2 & 21.2 & 8.9 & 0.0 & 0.0 & 0.0 & 0.0 \\
\midrule
5 X 15 & 3.2 & 2.0 & 1.3 & 0.2 & 28.1 & 28.1 & 7.2 & 3.3 & 0.0 & 0.0 & 0.0 & 0.0 \\
5 X 25 & 4.4 & 3.5 & 2.4 & 0.2 & 25.5 & 25.1 & 25.1 & 2.2 & 0.0 & 0.0 & 0.0 & 0.0 \\
5 X 50 & 4.1 & 3.5 & 2.7 & 0.6 & 14.2 & 14.1 & 14.1 & 14.1 & 0.0 & 0.0 & 0.0 & 0.0 \\
5 X 100 & 5.1 & 4.8 & 4.3 & 0.9 & 16.4 & 16.0 & 16.0 & 8.8 & 0.0 & 0.0 & 0.0 & 0.0 \\
\midrule
10 X 30 & 3.2 & 2.0 & 1.3 & 0.1 & 16.0 & 15.1 & 15.1 & 1.9 & 0.0 & 0.0 & 0.0 & 0.0 \\
10 X 50 & 3.0 & 2.5 & 1.9 & 0.2 & 8.7 & 7.7 & 7.7 & 4.9 & 0.0 & 0.0 & 0.0 & 0.0 \\
10 X 100 & 3.9 & 3.5 & 2.9 & 0.6 & 11.7 & 10.4 & 9.7 & 5.1 & 0.0 & 0.0 & 0.0 & 0.0 \\
10 X 200 & 5.5 & 5.3 & 5.1 & 2.2 & 12.2 & 11.9 & 11.9 & 7.3 & 0.0 & 0.0 & 0.0 & 0.4 \\
\midrule
25 X 75 & 3.1 & 2.3 & 1.8 & 0.3 & 11.0 & 11.0 & 7.9 & 1.5 & 0.0 & 0.0 & 0.0 & 0.0 \\
25 X 125 & 4.0 & 3.3 & 2.5 & 0.7 & 10.9 & 10.5 & 10.5 & 4.7 & 0.0 & 0.0 & 0.0 & 0.1 \\
25 X 250 & 4.4 & 3.9 & 3.6 & 2.0 & 9.9 & 9.5 & 9.5 & 5.6 & 0.0 & 0.0 & 0.0 & 0.5 \\
25 X 500 & 5.6 & 5.4 & 5.3 & 3.7 & 10.2 & 10.0 & 9.9 & 8.9 & 0.0 & 0.1 & 0.1 & 1.5 \\
\midrule
50 X 150 & 3.1 & 2.2 & 1.5 & 0.5 & 7.3 & 6.3 & 5.1 & 2.0 & 0.0 & 0.0 & 0.0 & 0.4 \\
50 X 250 & 4.4 & 3.5 & 3.2 & 1.7 & 9.2 & 8.8 & 8.8 & 5.2 & 0.0 & 0.0 & 0.1 & 0.8 \\
50 X 500 & 5.3 & 4.8 & 4.7 & 3.4 & 12.0 & 11.3 & 11.3 & 6.7 & 0.1 & 0.1 & 0.2 & 2.0 \\
50 X 1000 & 6.1 & 5.8 & 5.8 & 4.6 & 9.6 & 9.5 & 9.5 & 7.1 & 0.2 & 0.2 & 0.3 & 4.1 \\
\midrule
100 X 300 & 3.8 & 2.7 & 2.3 & 1.1 & 7.7 & 6.7 & 6.7 & 3.7 & 0.1 & 0.1 & 0.1 & 1.5 \\
100 X 500 & 4.4 & 3.5 & 3.2 & 2.3 & 6.9 & 5.7 & 5.7 & 4.0 & 0.1 & 0.2 & 0.2 & 2.8 \\
100 X 1000 & 5.5 & 4.9 & 4.9 & 4.0 & 9.8 & 9.2 & 9.2 & 6.2 & 0.3 & 0.4 & 0.5 & 6.0 \\
100 X 2000 & 6.4 & 6.1 & 6.1 & 5.1 & 10.2 & 9.8 & 9.8 & 7.3 & 0.7 & 0.8 & 1.1 & 12.7 \\
\midrule
200 X 600 & 3.4 & 2.3 & 2.0 & 1.1 & 4.9 & 4.0 & 3.9 & 2.2 & 0.3 & 0.4 & 0.5 & 5.9 \\
200 X 1000 & 4.4 & 3.6 & 3.4 & 2.8 & 7.3 & 6.7 & 6.7 & 4.0 & 0.6 & 0.6 & 0.8 & 9.6 \\
200 X 2000 & 5.5 & 5.0 & 5.0 & 4.4 & 8.1 & 7.5 & 7.4 & 6.2 & 1.2 & 1.4 & 1.6 & 20.0 \\
200 X 4000 & 6.3 & 6.0 & 6.0 & 5.4 & 9.5 & 9.3 & 9.3 & 7.2 & 2.5 & 3.1 & 3.6 & 43.4 \\
\midrule
400 X 1200 & 3.8 & 2.7 & 2.6 & 1.9 & 4.8 & 3.8 & 3.8 & 2.9 & 1.3 & 1.4 & 1.6 & 20.8 \\
400 X 2000 & 4.5 & 3.7 & 3.6 & 3.1 & 5.8 & 4.9 & 4.9 & 4.0 & 2.2 & 2.5 & 2.8 & 35.3 \\
400 X 4000 & 5.6 & 5.1 & 5.1 & 4.5 & 6.4 & 5.8 & 5.8 & 5.3 & 4.7 & 5.4 & 6.0 & 74.1 \\
400 X 8000 & 6.6 & 6.3 & 6.3 & 5.7 & 8.2 & 7.9 & 7.9 & 7.0 & 9.8 & 11.8 & 12.9 & 159.5 \\

 \bottomrule
\end{tabularx}
\end{table*}

In the case of general graphs, our tests show that the use of heuristic function $hs2$, selection of a subgraph with a minimal number of neighbors, is most beneficial with the average error between 0.7 - 8.2\%. According to Tables \ref{table:GenSelSub}, \ref{table:TreeSelSub} all the heuristics had  a better performance in case of a large ratio $m/n$ in case of the normalized error. This can be explained by the fact that in this case the cutoff of subgraphs was less likely which is a major source of error. In such cases, although the normalized error is small, generally less than 2\%, the absolute error is still significant. Our tests have shown that although $hs2$ gives better average results, it is less reliable than the balanced approach using $hs3$ which in the majority of the cases has a smaller average maximal error, which can go even up to 20\%. In case of trees we have a different behavior in several aspects. First, the decrease of average error with the increase of the number of demand nodes does not occur, since for this type of graphs, the use of the proposed algorithm is more likely to result in some subgraphs being cutoff.  For this type of graphs the balanced heuristic gives the best performance when average and maximal error are considered. One explanation for this is that in case of trees, the number of neighbors for each subgraph would vary much less, and without using a balanced approach the heuristic would frequently have a similar effect as a random search.

In our second group of tests we observe the effect of using different types of heuristics for node selection. For each of the graph types we have used the best performing heuristic for subgraph selection, from Tables \ref{table:GenSelSub},~\ref{table:TreeSelSub}, and combined them with different node selection heuristics. These results are presented in Tables \ref{table:GenSelNode},~\ref{table:TreeSelNode}, in which we present the same values as for the first type of heuristics. For general graphs we have found that the use of the heuristic  $hn1$ gives the best results for both average and maximal error. In case of trees, again we find that the balanced approach with heuristic $hn3$ finds the best quality solutions.

In the last group of tests we compare the effect of the two correction procedures, and the multiheuristic approach to the simple use of the greedy algorithm. This is done for both types of graphs, and we consider the average error and maximal error and total  execution time for the 40 test instances. As we can see from Tables \ref{table:GenCor}, \ref{table:TreeCor} the approaches are computationally very effective. Even in the case of the largest instances having 400 supply nodes and 8000 demands nodes the total execution time for 40 test instances is  10  and 7 seconds for general and tree graphs respectively when the greedy algorithm is considered. The increase in the execution time of the non-located and combined corrections was dependent on the average number of connections in the graphs. In case of general graphs the increase of the  calculation time of the non-located and combined correction  was approximately 50\% and 150\%, while in the case of trees it was  20\% and  50\% of the calculation time of the greedy algorithm. The execution time of the multiheuristic approach was around 12 times of the combined heuristic correction, which is expected since 12 such runs performed. The  level of improvement of the average error was significant for both  of the correction methods, and managed to improve it in almost all the sets of instances.  In case of the average maximal error, the performance of the  corrections was dependent on the graph size, and as it increases so does the level of improvement. For several graph sizes the corrections did not manage to improve this value which indicated that strong locally optimal solutions exists which can not be escaped using this procedure.

The  use of the multiheuristic approach produced a great improvement in the quality of found solutions. The average error was less than 1\% in 40\% of the graph sizes, it never exceeding 5.7\%, while having an error  greater than 4\% in less than 20\% of the graph sizes.  The use of this approach has proven to be very robust and has given an improvement for the maximal error in almost all the cases. Out of the 64 test sizes for only two of them it gave a maximal error greater than 10\%.

\section{Conclusion}

In this paper we presented a method for  solving  the problem of maximal partitioning of graphs with supply and demand, with a special focus on solving large scale problem instances. The main goal of this work was to provide some experimental results for the problem as most of the previous works were more of a theoretical nature. Test data sets and the source code for the implemented algorithm are publicly available.  The proposed method consists of a greedy algorithm based on two separate heuristics, and two correction procedures. We have presented several potential heuristic functions for the two stages of the algorithm and analyzed their effectiveness on sparse general graphs and trees. The calculation time for the proposed correction procedures is similar to the one of the greedy algorithm, while  managing to considerably improve the initially acquired solutions.

Our experiments have shown that the performance of  competing heuristics significantly differs depending on the graph properties. Due to the fact that the correction procedure, to a large extent, behaves like a hill climbing  method, the combination of different heuristics with it produce very good results. This type of multiheuristic approach manages to produce high quality average solutions while at the same time being very robust. In the conducted experiments it has been shown that even for very large graphs consisting of more than 8000 nodes, good quality solutions can be found within  seconds.


%
%


\end{document}